\documentclass{article}

\PassOptionsToPackage{numbers, compress}{natbib}
\usepackage[preprint]{neurips_2024}
\PassOptionsToPackage{square,numbers}{natbib}
\usepackage{graphicx}
\usepackage{multirow}
\usepackage{amsmath,amssymb,amsfonts}
\usepackage{mathtools}
\usepackage{amsthm}
\usepackage{mathrsfs}
\usepackage[title]{appendix}
\usepackage[table]{xcolor}
\usepackage{textcomp}
\usepackage{manyfoot}
\usepackage{booktabs}
\usepackage{algorithm}
\usepackage{algorithmicx}
\usepackage{algpseudocode}
\usepackage{listings}
\usepackage{verbatim}
\usepackage{subcaption}
\usepackage{makecell}
\usepackage{adjustbox}
\usepackage{comment}
\usepackage{color,soul}
\usepackage{hyperref}
\usepackage{url}
\usepackage{tabularray}
\usepackage{lmodern}

\usepackage[utf8]{inputenc} % allow utf-8 input
\usepackage[T1]{fontenc}    % use 8-bit T1 fonts
\usepackage{hyperref}       % hyperlinks
\usepackage{url}            % simple URL typesetting
\usepackage{booktabs}       % professional-quality tables
\usepackage{amsfonts}       % blackboard math symbols
\usepackage{nicefrac}       % compact symbols for 1/2, etc.
\usepackage{microtype}      % microtypography
\usepackage{xcolor}         % colors

\title{Automatic Differential Diagnosis using Transformer-Based Multi-Label Sequence Classification}

\author{%
  Abu Adnan Sadi\thanks{Corresponding author} \\
  Department of Electrical and Computer\\ 
  Engineering, North South University,\\
  Dhaka, 1229, Bangladesh\\
  \texttt{abu.sadi05@northsouth.edu} \\
  \And
  Mohammad Ashrafuzzaman Khan\\
  Department of Electrical and Computer\\ 
  Engineering, North South University,\\
  Dhaka, 1229, Bangladesh\\
  \texttt{mohammad.khan02@northsouth.edu} \\
  \AND
  Lubaba Binte Saber \\
  Sir Salimullah Medical College, \\
  Dhaka, 1100, Bangladesh \\
  \texttt{lbsrefa@gmail.com} \\
}

\begin{document}

\maketitle

\begin{abstract}
  As the field of artificial intelligence progresses, assistive technologies are becoming more widely used across all industries. The healthcare industry is no different, with numerous studies being done to develop assistive tools for healthcare professionals. Automatic diagnostic systems are one such beneficial tool that can assist with a variety of tasks, including collecting patient information, analyzing test results, and diagnosing patients. However, the idea of developing systems that can provide a differential diagnosis has been largely overlooked in most of these research studies. In this study, we propose a transformer-based approach for providing differential diagnoses based on a patient's age, sex, medical history, and symptoms. We use the DDXPlus dataset, which provides differential diagnosis information for patients based on 49 disease types. Firstly, we propose a method to process the tabular patient data from the dataset and engineer them into patient reports to make them suitable for our research. In addition, we introduce two data modification modules to diversify the training data and consequently improve the robustness of the models. We approach the task as a multi-label classification problem and conduct extensive experiments using four transformer models. All the models displayed promising results by achieving over 97\% F1 score on the held-out test set. Moreover, we design additional behavioral tests to get a broader understanding of the models. In particular, for one of our test cases, we prepared a custom test set of 100 samples with the assistance of a doctor. The results on the custom set showed that our proposed data modification modules improved the model's generalization capabilities. We hope our findings will provide future researchers with valuable insights and inspire them to develop reliable systems for automatic differential diagnosis. 
\end{abstract}

\paragraph{Keywords:}Natural Language Processing, Healthcare, Transformers, Multi-label Classification, Differential Diagnosis, Behavioral Testing.

%%%%%%%%%%%%%%%%%%%%%%%%%%%%%%%%%%%%%%%%%%%%%%%%%%%%%%%%%%%%%%%
\section{Introduction}

In a clinical setting, doctors have to perform the task of differential diagnosis on a regular basis \cite{first2024dsm}. During their consultation with doctors, patients generally provide information such as their symptoms, how severe the symptoms are, and how long they have been experiencing these symptoms. Based on these pieces of information, a doctor then has to make a differential diagnosis, which is a list of possible medical conditions that may cause such symptoms. Sometimes, the doctor may also have to review the medical history of the patient to narrow down the list of possible causes. After a differential diagnosis is made, doctors usually order additional tests to confirm the final diagnosis. Differential diagnosis plays an important role in identifying what a patient is suffering from, allowing the doctor to provide suitable treatments.

Techniques based on artificial intelligence (AI) have become increasingly effective tools for transforming healthcare. Researchers have made great strides in the development of automatic disease diagnosis (ADD) systems by applying machine learning and deep learning-based techniques. When it comes to diagnosing diseases from medical images, deep convolutional neural networks (DCNNs) have greatly improved the performance of ADD systems \cite{yadav2019deep, marques2020automated, raj2020optimal}. Researchers have also used techniques based on natural language processing (NLP) for analyzing medical texts and performing ADD tasks \cite{Byrd2014AutomaticIO, liang2019evaluation, feller2018using}. In recent times, the proposal of the transformer architecture \cite{vaswani2017attention} has completely revolutionized the field of NLP. The capability of transformer models to understand the context of natural language has made it the state-of-the-art technology in NLP. As a result, it has gained huge popularity among researchers in the medical domain. Several studies have shown how transformer models can be utilized to perform ADD tasks using both medical texts \cite{rasmy2021med, prakash2021rarebert, rao2022explainable} and medical imaging\cite{he2023transformers}, achieving exceptional results in both areas.

Although a lot of research is being done to enhance transformer-based methods for disease diagnosis, the majority of these disease classification models are made to produce a single diagnosis based on a given medical text data. The concept of differential diagnosis is still widely under-represented in the healthcare NLP domain. The lack of differential diagnosis in existing NLP-based methods is a severe problem since it omits a crucial step in the medical diagnostic process. It is essential to have multiple (both common and rare) diseases in a differential diagnosis, as it improves the likelihood of accurate identification of the actual disease \cite{Jain2017}. 

\begin{figure}
    \centering         
    \includegraphics[width=\linewidth]{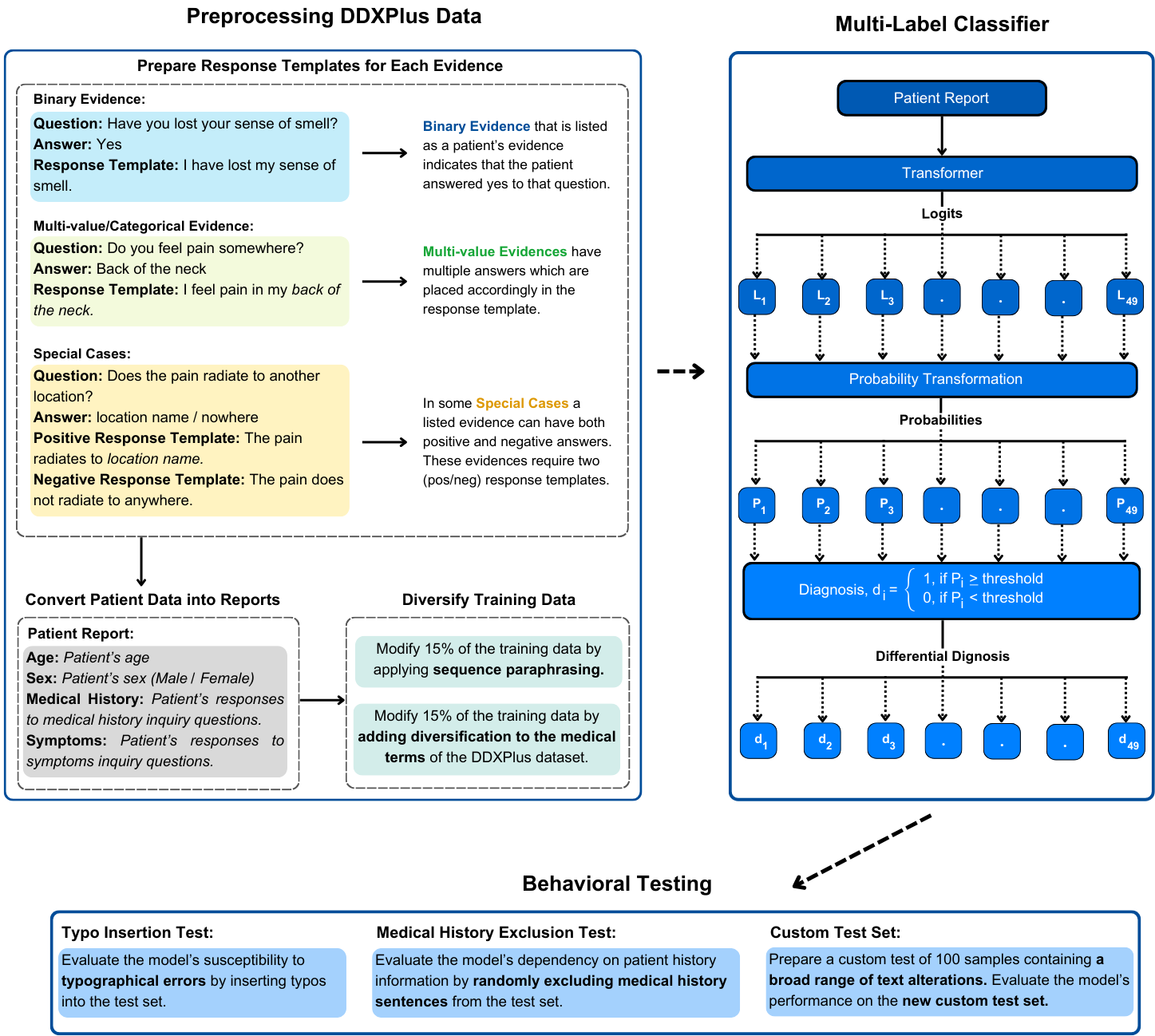}
    \caption{\textbf{Schematic diagram of the proposed approach for making automatic differential diagnosis.}}
    \label{fig:schematic}
\end{figure}

In our study, we address this research gap by demonstrating how transformer-based architectures can be utilized to perform differential diagnosis. In order to achieve our goal, we treat the task of performing a differential diagnosis as a multi-label classification (MLC) problem. Given a sequence of text containing a patient's age, sex, medical history, and symptoms, the transformer model will output a list of potential diseases that can cause such symptoms. We have used the publicly available DDXPlus dataset \cite{fansi2022ddxplus} to fine-tune our model for the task of differential diagnosis. The dataset includes 49 disease categories associated with medical conditions like sore throats, coughs, and breathing difficulties. We apply several dataset processing techniques to convert the patient samples of the DDXPlus dataset into textual patient reports that can used for fine-tuning a transformer model. Moreover, we propose two data modification modules to apply sequence paraphrasing and diversify medical terms in the training data to improve the robustness of the models. Lastly, by taking inspiration from \cite{ribeiro2020beyond}, we evaluate the robustness and limitations of our trained models by designing additional test cases. We design three behavioral tests to evaluate the performance of our proposed approach further. The illustration in \textbf{Fig \ref{fig:schematic}} provides a simplified overview of the primary contributions of this paper.

%%%%%%%%%%%%%%%%%%%%%%%%%%%%%%%%%%%%%%%%%%%%%%%%%%%%%%%%%%%%%%%
\section{Related Works} 
Based on our research, we categorize this section into the following groups:

\begin{itemize}
    \item \textbf{NLP for Medical Diagnoses.} For many years, researchers have been utilizing natural language processing to identify and extract clinical information from medical text, such as electronic health records(EHRs)\cite{koleck2019natural}. The recent development of transformer architectures \cite{vaswani2017attention} and large-scale public datasets such as MIMIC-III \cite{johnson2016mimic} and MIMIC-IV \cite{johnson2023mimic} have allowed researchers significant progress in developing NLP-based patient healthcare systems. Several studies have proposed novel transformer-based models that are pre-trained using EHRs \cite{rasmy2021med, li2020behrt}, which can be used for disease predictions. In another study \cite{prakash2021rarebert}, the authors proposed a novel BERT architecture named RareBERT for improving rare disease diagnosis. They tested their architecture to identify a rare genetic disease called X-linked Hypophosphatemia using administrative claims data. Several studies have also employed the sequence classification approach for performing an automatic diagnosis. For instance, in \cite{murarka2021classification}, the authors used BERT and RoBERTa-based classifiers to identify signs of mental illness from social media posts. The identification of mental illness was treated as a multi-class problem where the models classified the posts into one of the five illness categories. In another work \cite{chen2022diaformer}, the authors proposed a novel transformer-based decoder–encoder architecture, where the encoder is used for performing disease diagnosis as a sequence classification task. In addition, several studies have done automatic text classification of radiology reports for identifying medical conditions, including intracranial hemorrhage \cite{jnawali2019automatic}, Multiple sclerosis (MS) \cite{rietberg2023accurate}, ischemic stroke \cite{ong2020machine}, and tinnitus \cite{li2022automatic}. A lot of the aforementioned studies have treated the task of medical diagnosis as a binary or multi-class classification problem, where a given text is classified as one of the classes present within the training dataset. In this study, we treat the task of performing medical diagnosis as a multi-label text classification problem, allowing us to obtain differential diagnoses.

    \item \textbf{Multi-Label Classification of Texts.} Multi-label classification (MLC) is primarily used in natural language processing for performing the task of Document classification \cite{rubin2012statistical, lenc-kral-2017, haghighian2022patentnet}. Several studies have also applied MLC to classify clinical documents \cite{blanco2019multi, liu-2021-effective, baumel2018multi, trigueros2022explainable}. For instance, in their work \cite{blanco2019multi}, the authors proposed a deep-learning-based MLC system for classifying EHRs to their corresponding International Classification of Diseases (ICD) codes. In their work \cite{liu-2021-effective}, the authors developed an effective convolutional attention network for classifying long clinical documents. They used discharge summaries from the MIMIC-III to predict their ICD-9 codes. Instead of using large clinical documents (EHRs), in this study, we use concise clinical texts containing salient patient information, such as age, sex, medical history, and symptoms. We then apply a multi-label sequence classification technique to provide a differential diagnosis based on the given information.
    
    \item \textbf{Multi-Label Diagnosis.} Prior studies have widely used multi-label disease diagnosis approaches in the computer vision domain. Several studies have used eye fundus images to detect retinal diseases using both CNN \cite{wang2020multi} and transformer-based \cite{rodriguez2022multi} models. In another study, the authors applied an MLC method to detect breast cancer from mammogram images. Similarly, multiple studies have applied multi-label classification techniques to diagnose diseases from chest x-ray images \cite{guan2020multi, chen2020label, xiao2023delving}. In contrast to computer vision-based techniques, very few studies have used multi-label classification approaches for disease diagnosis in the NLP domain. In \cite{li2020behrt}, the author used patients' past EHR records to predict their future diagnoses by applying MLC. In another work \cite{chaichulee2022multi}, the authors applied an MLC approach using BERT-based models to obtain symptom terms from free-text drug allergy descriptions. Most of these prior works employ MLC toward the final phase of the diagnostic process, where they determine the medical conditions a patient is suffering from. In contrast, our study focuses on the earlier stages of the diagnostic assessment, where we provide a list of potential diseases that might cause the symptoms the patient is experiencing. Using this differential diagnosis, a doctor can then order additional tests to reach the final diagnosis.
    
    \item \textbf{Automatic Differential Diagnosis.} There are also a very limited amount of research studies that use text analysis to offer differential diagnoses according to patient symptoms. The authors of \cite{fansi2022ddxplus} used their proposed DDXPlus on existing models: AARLC \cite{yuan2024efficient} and BASD \cite{luo2021knowledge}, both of which rely on interacting with patients symptom acquisition. The AARLC model utilized a reinforcement learning (RL) module to inquire about symptoms. In comparison, the BASD model used a prediction module, which predicts whether to continue or stop the inquiry process and what symptoms to ask about next. Both the AARLC and BASD models employed a classifier network to provide the final disease prediction. In contrast to the proposed method of \cite{fansi2022ddxplus}, we aim to simplify the task of providing a differential diagnosis by using a single text classification module. Instead of developing a patient inquiry system, we simply ask for a short description of the patient's medical history and symptoms to provide the diagnosis. To the best of our knowledge, this is the first study that tackles the challenge of providing automatic differential diagnosis using a multi-label classification strategy.
\end{itemize}

%%%%%%%%%%%%%%%%%%%%%%%%%%%%%%%%%%%%%%%%%%%%%%%%%%%%%%%%%%%%%%%
\section{Methodology}
\subsection{Dataset Processing}\label{sec:dataset_processing}
The DDXPlus dataset is a large-scale synthetic benchmark dataset that includes around 1.3 million patient data with 110 symptoms, 113 antecedents (patient medical history), and 49 diseases. The dataset includes a combination of binary, multi-choice, and categorical symptoms and antecedents. It was created using a commercial automatic diagnosis system and a proprietary medical knowledge base. The diseases included in the dataset are mostly related to medical problems such as cough, sore throat, or breathing issues. 

The dataset was primarily designed to create efficient evidence-collection systems, in which an AI agent would ask a patient about their symptoms and medical history and then make a differential diagnosis based on the patient's responses. As a result, the dataset contains a total of 223 (symptoms and antecedents) questions, which can be answered by the patients with a simple Yes/No (binary) or with a multi-choice or categorical value (such as location of pain, intensity of pain, color of rash, etc.). The authors characterized these questions as evidences, and each patient data includes a list of evidences (with answers) experienced by the patient. Additional information for each patient includes their age and sex, along with a differential diagnosis based on the evidence. The patient data is stored in a tabular format in the dataset. Please refer to the DDXPlus dataset paper \cite{fansi2022ddxplus} to get a clear understanding of the dataset structure.

\begin{figure*}
    \centering         
    \includegraphics[width=\textwidth]{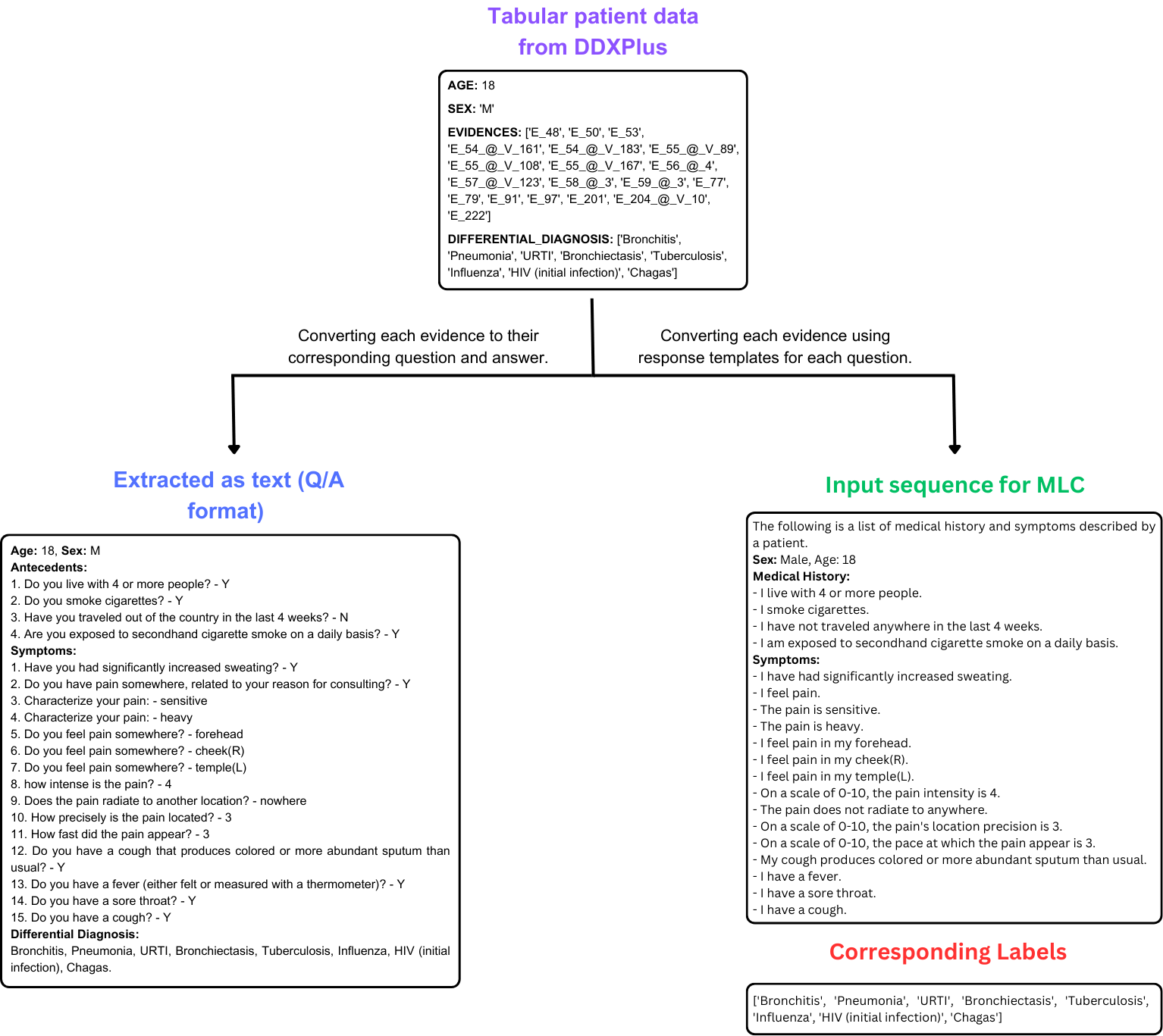}
    \caption{\textbf{Conversion process of DDXPlus patient samples.} Tabular data can be directly extracted in a question-answer format using evidence values (bottom left). Our approach for converting the tabular data into text reports using response (answer) templates for each question (bottom right).}
    \label{fig:ddxplus_sample}
\end{figure*}

Firstly, we extracted the tabular data and converted it into a text format (see \textbf{Fig \ref{fig:ddxplus_sample}}). As can be seen from the figure, since the evidences are stored as questions in the dataset,  the data can directly be extracted in a question-answer format. However, this question-answer format is not suitable for our end goal, as we want to design a system where a patient describes their medical history and symptoms and based on which the model provides a differential diagnosis. As a result, we next prepared short sentence templates that act as a response to each of the 223 unique questions present in the dataset. A few examples of how we created these response templates are shown in \textbf{Fig \ref{fig:schematic}}. To make the responses more reflective of how a real patient might explain their symptoms, they were written in the first person. Given the patient's age, sex, and a list of evidences, our code is designed to construct a block of text (patient report) that portrays a patient reporting their own symptoms (see \textbf{Fig \ref{fig:ddxplus_sample}}). Finally, a new dataset consisting of these input sequences and their corresponding labels was used to train the transformer models for multi-label classification.

The primary goal behind converting the DDXPlus dataset into a dataset of patient reports and using transformer-based architectures is to develop robust models that can recognize the contextual patterns within the text. If we had designed an evidence collection system using the raw dataset, then the algorithm would only be able to ask questions that are present within the dataset and use the predetermined answer choices to make predictions. Furthermore,  if we had used the initially extracted question-answer format text for training, then the model might become biased and only be able to perform well as a classifier given that the input is also in a similar question-answer format. This would have been counter-intuitive to the goal of this study. On the other hand, using a block of text that represents a patient describing their symptoms and medical history greatly simplifies the classification task. Here, a user of this model would only have to provide a list of the symptoms and medical history, along with the age and gender of the patient, to receive a differential diagnosis. Even if the input sequence has a slightly different structure than the sequences used during training, due to the robustness and the capability of transformer models to understand context, the model would be able to provide a meaningful diagnosis.

\subsection{Data Modification Modules}\label{sec:data_modification}
As mentioned in \textbf{section \ref{sec:dataset_processing}}, the DDXPlus dataset contains a total of 223 questions related to a patient's symptoms and medical history (antecedents). We then prepared response templates for each of these questions to prepare our dataset for the MLC task. Therefore, each patient in the classification dataset contains some combination of these response templates that describe their medical history and symptoms. Due to the limited number of response sentences, the training dataset lacks some lexical variations. This lack of variation may negatively affect the performance of the model on unseen data that use different sentence structures and vocabularies. 

In addition, the DDXPlus dataset contains a number of medical terms that are particularly related to human anatomy. These terms are used to indicate the body parts in which a patient might have pain, rash, or swelling. However, these medical terms are not exhaustive. For example, a patient may not be aware of the anatomical term of a body part and may indicate the location of discomfort using a more general term or phrase. For instance, to describe pain in the `Epigastric' (an anatomical term used in DDXPlus) region, a patient might say that they feel pain in the `upper abdomen' region. In another scenario, a doctor might use a widely accepted synonym for a particular medical term present in the DDXPlus Dataset. For example, a doctor might use synonym terms such as `Antecardium' or `Epigastrium' to indicate the `Epigastric' region. Therefore, in order to make the model more capable of understanding the connections between these related terminologies, it is necessary to introduce diversity to medical terms present in the DDXPlus dataset.

To mitigate these issues, we incorporate two types of data modification modules, 1) Sequence Paraphrasing (SP) and 2) Medical Term Diversification (MTD), into our pipeline to diversify our training data.

\subsubsection{Sequence Paraphrasing}\label{sec:seq_paraphrasing}
As we can see from \textbf{Fig \ref{fig:ddxplus_sample}}, an input sequence contains lists of sentences that describe a patient's medical history and symptoms. In order to paraphrase these sentences, we employ the following two steps-

\begin{enumerate}
    \item \textbf{Sentence Selection:} Let's consider that the input sequence $S$ contains $N$ number of sentences that are related to a patient's medical history or symptoms. We randomly select a subset of sentences $\hat{S} \subset S$ in order to apply paraphrasing. In our study, we randomly selected 40\% ($0.4 * N$) of the sentences for paraphrasing. We only selected a subset of sentences to paraphrase because we wanted to add variety to the text without compromising its primary context.
    \item \textbf{Paraphrasing and Replacement:} The selected sentences are paraphrased using a paraphrasing tool. For our study, we use the OpenAI API's GPT-3.5 Turbo model\footnote{\url{https://platform.openai.com/docs/models/gpt-3-5-turbo}} as a paraphrasing assistant. We utilize such a powerful model because our goal was to preserve the semantic meaning of the original sentences after paraphrasing. This is something a lot of smaller models fail to achieve. Once the selected sentences are paraphrased, they are then substituted back into the input sequence, replacing the original sentences.
\end{enumerate}

\subsubsection{Medical Term Diversification}
The DDXPlus dataset contains a little over 160 anatomical terms or phrases that are used to indicate different parts of a patient's body. In order to find related terms for these medical terms, we primarily use the SNOMED-CT \cite{donnelly2006snomed} terminology browser. SNOMED-CT is one of the most comprehensive clinical terminology datasets, which was created to assist medical doctors and researchers in standardizing the usage of clinical terms in electronic health records (EHRs). The SNOMED CT includes synonyms for every medical term that are both preferred and widely acceptable. Thus, we utilize this SNOMED-CT terminology browser to find synonyms for the anatomical terms present in the DDXPlus dataset. In addition, the SNOMED-CT database contains a parent-child relationship for each medical term, which we further utilized to find more generalized terms that are related to the anatomical terms in the DDXPlus dataset. 

For example, the term `hypochondrium(R)' is a term in the DDXPlus dataset that indicates the right hypochondrium of a patient. From SNOMED-CT, we can find the term `hypochondriac region' is a preferred synonym for hypochondrium. We can also find from the parent-child relationship that the Hypochondrium is a structure of the upper abdominal quadrant. Therefore, we include `right side of upper abdominal quadrant' and `right hypochondriac region' as related terms for the `hypochondrium(R)' term. In addition, we also expand terms that contain a left or right (R or L) indicator. So, we expand `hypochondrium(R)' to `right hypochondrium' and include it as a related term. As a result, the final list of related terms for `hypochondrium(R)' would contain the following: right hypochondriac region, right side of upper abdominal quadrant, right side of the upper abdomen, and right hypochondrium. 

However, there were some medical terms for which we did not collect any related terms, as it was simply not necessary. For instance, terms such as mouth, nose, knee, and thigh are already very generalized and widely used. In such cases, we either simply did not include any related terms or expanded the terms that contained a left or right (R or L) indicator and kept the expanded form as a related term.

Overall, we collected 1 to 4 related terms for each of the anatomical terms present in the dataset. We verified all the collected terminologies by consulting with a doctor. When applying the modifications, we first select a proportion of samples from the training set. In the selected samples, we replace all medical terms from the DDXPlus dataset with one of their corresponding related terms.

\subsection{Multi-Label Sequence Classificaiton}
\subsubsection{Model Training}
Let $R = \{r_1, r_2,...., r_N\}$ be a set of patient reports containing textual information related to the patient's age, sex, medical history, and symptoms. Our goal is to obtain a differential diagnosis for a patient by classifying their report $r_i \in R$ into a subset of related disease labels $d_i \subset L$, where $L = \{l_1, l_2,...., l_m\}$ is a set of all labels. We encode the ground-truth labels using the one-hot encoding method.

Next, we tokenize the input text (patient reports), which adds a classification token [CLS] (\textless s\textgreater token for RoBERTa) to the start of the input text. The hidden state representation of the [CLS] token obtained from the last encoder layer is later used for classification. Let $H_{CLS}$ be the hidden state representation of the [CLS] token obtained from the last encoder layer of the transformer. Then, the output from the classification layer is obtained as follows:

\begin{equation}
    \begin{aligned}
    & h_l = FC_1(H_{CLS}) \\
    & h_{nl} = tanh(h_l)\\ 
    & W_C = FC_2(h_{nl})
\end{aligned}
\label{eq:model_output}
\end{equation}

Here, $H_{CLS}$ is first fed into a fully connected layer $FC_1$ to obtain a linear representation $h_l \in \mathbb{R}^{batch\_size \times \lvert H_{CLS}\lvert}$, which then goes through a tanh() activation function that applies non-linear transformation. Finally, the transformed output $h_{nl}$ is fed through a second fully connected layer $FC_2$ to obtain the final classifier output $W_C \in \mathbb{R}^{batch\_size \times \lvert L \lvert}$. 

The model then calculates the loss value using $W_C$, which is used to calculate the gradients during backpropagation. We use the Binary Cross Entropy with Logits (BCEL) loss as a loss function, which is suitable for multi-label classification. BCEL loss is a binary cross-entropy loss with a sigmoid function. The sigmoid function converts the logits obtained from $W_C$ into probability values ranging between 0 and 1. The formula for the sigmoid function is as follows:

\begin{equation}
    \sigma(x) = \frac{1}{1 + e^{-x}}
\label{eq_sigmoid}
\end{equation}

Now, the formula for BCEL loss can be expressed as:

\begin{equation}
 \begin{aligned}
    & L_{BCEL} = -\frac{1}{N} \sum_{i=1}^{N} \sum_{j=1}^{M} \: [y_{ij} \: log(p_{ij}) + (1-y_{ij}) \: log(1-p_{ij})] \\
    & where,\:\: p_{ij} = \sigma(\hat{y}_{ij})
\end{aligned} 
\label{eq_bcel}
\end{equation}

Here, $N$ is the total number of samples and $M$ is the total number of labels. For the $i^{th}$ sample, $y_{ij}$ and $p_{ij}$ are the true label (0 or 1) and predicted probability for the $j^{th}$ label, respectively. Similarly, $\hat{y}_{ij}$ is the logit value obtained from the classifier output $W_C$. This BCEL loss value is used to update the model weights during training.

\subsubsection{Model Evaluation} \label{sec:model_evaluation}
During the evaluation process, we first obtain the logits from $W_C$, which are then turned into probability values using \textbf{equation \ref{eq_sigmoid}}. From this point forward, we will refer to these probability values as model confidence scores. Next, we define a confidence threshold value, $T_{conf}=0.5$. The confidence score for each label is then converted to 1 if it is greater than or equal to $T_{conf}$ and 0 if it is not. Finally, the disease labels for which the confidence score is set to 1 are extracted to obtain the differential diagnosis.

\subsection{Behavioral Testing} \label{sec:behavioral_test}
Like most previous studies, we also used a traditional held-out test set to evaluate the performance of our models after training. However, we wanted to put our models through further tests to determine their robustness and limitations. Inspired by the work of \cite{ribeiro2020beyond}, we designed additional test cases that are suitable for better understanding the capabilities of the models for the differential diagnosis task. The authors of \cite{ribeiro2020beyond} demonstrated how different test cases can expose flaws and limitations in models that otherwise perform almost as well as humans on traditional benchmarks. One could design new test cases by altering an existing test set or by starting from scratch. In our study, we explored both such approaches.

\subsubsection{Typo Insertion}
Typographical errors (typos) in the input text are very frequently encountered in natural language processing tasks. In a similar fashion, any consumer-focused natural language tool that is designed to provide differential diagnoses will also receive input text that may contain multiple typos. Therefore, it is necessary to understand how susceptible our models are to texts containing typographical errors. We prepared a separate test set by inserting different kinds of typos in all the samples of our held-out test set. We specifically targeted sentences that were part of the medical history and symptoms sections of the input text. We randomly selected 50\% of these sentences to insert typos. Each short sentence was inserted with at least one typo, whereas longer sentences contain multiple. While it is rare for a real-world text to have such a high amount of typos, we intentionally made our test difficult to account for the extreme cases. 

We inserted the following five kinds of typos randomly into our test set:
\begin{enumerate}
    \item \textbf{Replace with Adjacent Character:} Randomly replace one of the characters of a word with an adjacent character from the keyboard layout. For example, $Sample \rightarrow Samplr$.
    \item \textbf{Add Extra Adjacent Character:} Randomly select one character from a word and then add an extra adjacent character next to the selected character. For example, $Sample \rightarrow Sasmple$.
    \item \textbf{Swap Characters:} Randomly select one of the characters of a word and then swap it with the next consecutive character in that word. For example, $Sample \rightarrow Sapmle$.
    \item \textbf{Skip Characters:} Randomly skips a character from a selected word. For example, $Sample \rightarrow Smple$.
    \item \textbf{Repeat Characters:} Randomly repeat one of the characters of a word. For example, $Sample \rightarrow Samplle$.
\end{enumerate}

\subsubsection{Medical History Exclusion}
There is always a possibility that a patient might forget to mention all relevant information regarding their medical history. Therefore, it is necessary to identify how much it will affect a model to make a differential diagnosis in such a scenario. For that reason, we designed another test set where we randomly excluded a proportion of sentences from the medical history section of the held-out test set samples. We randomly defined the proportion from a predefined set of options: 50\%, 60\%, 70\%, 80\%, 90\%, or 100\%. As a result, there were some samples where the medical history text is completely removed. This enabled us to understand better how our models rely on information about a patient's medical history.

\subsubsection{Custom Test Set}
In this behavioral test, we prepared a new custom test set of 100 samples. We designed this test set from scratch by using the test samples from the DDXPlus dataset as a starting point. The patients (samples) in this custom dataset are associated with the same 49 diseases (pathologies) as the DDXPlus dataset. We include both a differential diagnosis and a ground-truth diagnosis for each patient. We ensured that the dataset contained at least two samples (based on ground-truth diagnosis) for each of the 49 diseases.

\begin{figure*}
    \centering         
    \includegraphics[width=\textwidth]{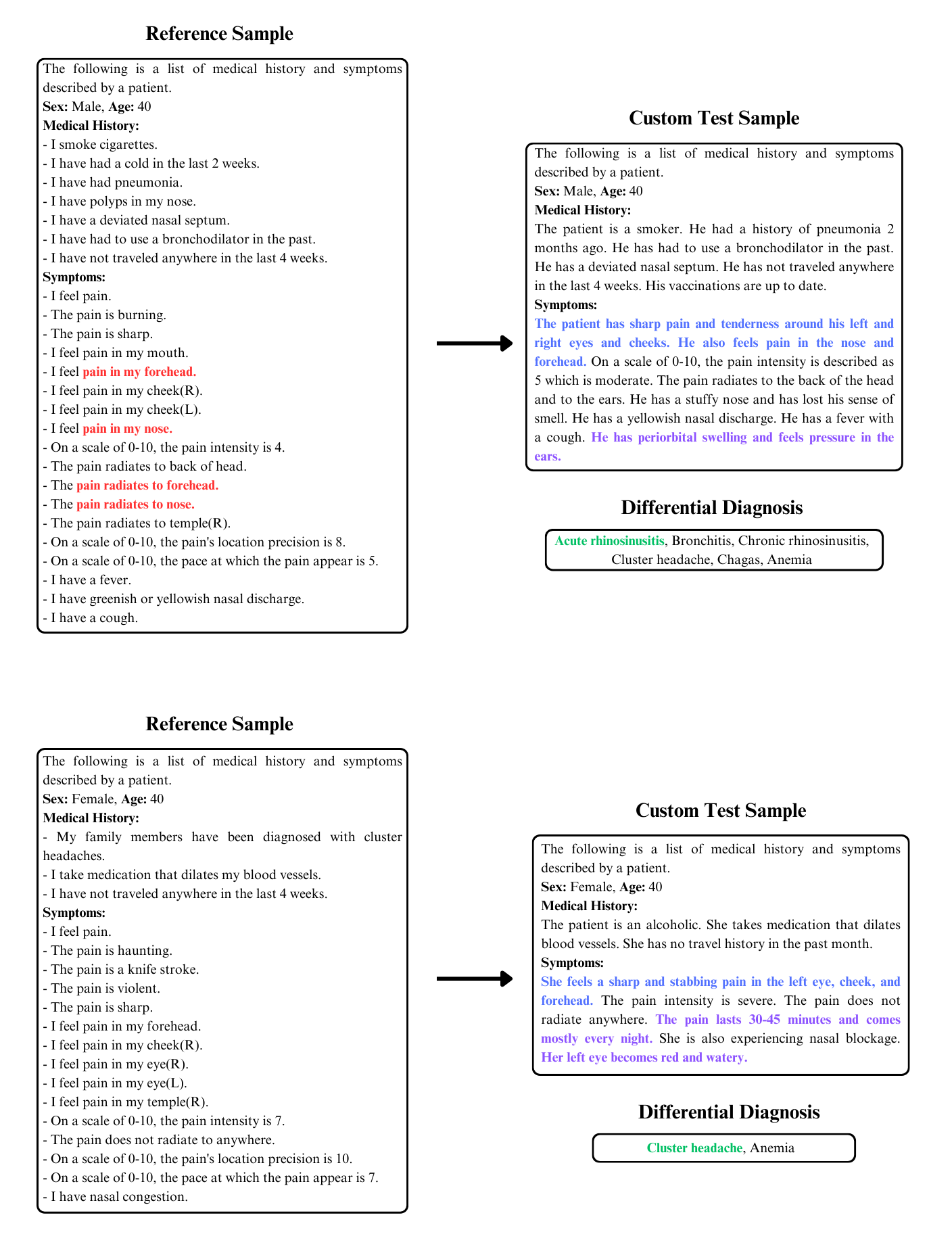}
    \caption{\textbf{Samples from the custom test set.} The purple-colored text indicates newly added symptoms, the blue-colored text indicates complex sentences containing several pieces of information, the red-colored text indicates repetitive information within the reference text, and the green-colored text indicates the ground-truth diagnosis for that patient.}
    \label{fig:custom_test_samples}
\end{figure*}

We created this custom test set because it enabled us to evaluate our models across a broad range of text alterations. This was a task that could not be accomplished systematically with code. Our goal was to make this new test set extremely challenging by introducing perturbations that were not present in the original training data. We sought the assistance of a medical doctor in developing this new test set. \textbf{Fig \ref{fig:custom_test_samples}} showcases some samples from the new test set. It is worth mentioning that we considered multiple similar reference samples from the DDXPlus dataset before creating a new test sample. For illustration purposes, we only show a single reference sample in the figure.

The following is a list of text perturbations that we applied to this new custom test set:

\begin{enumerate}
    \item All the samples in the custom test set were written in the third person (see \textbf{Fig \ref{fig:custom_test_samples}}). Similar to how a doctor or a nurse would write down a patient's information. In contrast, all the samples in our training set were written in the first person, simulating how a patient might describe their symptoms.
    
    \item All the samples in the custom test set were written in a paragraph format (see \textbf{Fig \ref{fig:custom_test_samples}}). In contrast, all the samples in our training set were organized in a list structure, with each patient providing a complete list of their medical history and symptoms.
    
    \item We also introduced new symptoms in some of the custom samples. Usually, this type of information was not part of the training data. We included such information to test how out-of-distribution data affects the performance of the models. \textbf{Fig \ref{fig:custom_test_samples}} demonstrates how some of this new information was added (marked in purple).
    
    \item We also paraphrased a lot of the information in the new test set in comparison to the sentences in the training dataset. While we did use a paraphrasing module to diversify our training data (as described in \textbf{section \ref{sec:seq_paraphrasing}}), that module only paraphrases a proportion of sentences where each sentence would only describe a single symptom or medical history. Whereas we apply much more extreme modifications to describe multiple pieces of information in a single sentence using comma separation and conjunctions. From \textbf{Fig \ref{fig:custom_test_samples}}, we can see some examples of such complex sentences (marked in blue). Furthermore, we often used similar and simpler phrases in place of complex medical terminology. 
    
    \item Lastly, We ensured that all the samples were kept concise and did not contain excessive information. As the DDXPlus dataset was synthesized using a medical knowledge base, some samples often contain repetitive information. For example, we can see from \textbf{Fig \ref{fig:custom_test_samples}}, the reference text mentions the patient feels pain in the forehead and nose (marked as red). However, it also mentions the pain radiates to the forehead and nose. This can be seen as a little redundant because stating that the pain radiates to a certain region implies that the patient feels pain in that region. In our custom test set, we tend to avoid such redundant information.
\end{enumerate}

%%%%%%%%%%%%%%%%%%%%%%%%%%%%%%%%%%%%%%%%%%%%%%%%%%%%%%%%%%%%%%%
\section{Experimental Settings}

\subsection{Dataset Distribution} \label{sec:dataset_dist}
The DDXPlus dataset contains approximately 1.3 million samples (patients), which are divided into training, validation, and test sets with an 80:10:10 split. However, we were unable to train our models on such a large number of samples due to a lack of computational resources. Therefore, we selected a subset of samples from each of the training, validation, and test sets of the DDXPlus dataset. In addition to the differential diagnosis, each patient in the dataset has a ground truth pathology, which is the actual disease the patient is suffering from. For the training set, we selected 1000 samples from each of the 49 diseases present in the dataset. Similarly, we selected 100 samples from each of the 49 disease categories for the validation and test sets. 

We selected an equal number of patients based on their ground truth pathology in order to avoid an imbalance in our training data. When considering the patient's ground truth pathologies, the DDXPlus dataset is extremely imbalanced. In particular, the pathologies Ebola and Bronchiolitis have less than 1000 samples available for the training set. As a result, our training dataset contained a total of 47979 samples, with 718 samples for Ebola and 261 samples for Bronchiolitis. Similarly, our validation dataset contained 4818 samples, with 90 samples for Ebola and 28 samples for Bronchiolitis. Our test dataset had 4836 samples, with only 36 samples for Bronchiolitis.

Lastly, when applying the data modifications mentioned later in \textbf{section \ref{sec:data_modification}}, we randomly selected 15\% of the training set to apply sequence paraphrasing and another 15\% to apply medical term diversification. The rest of the 70\% samples of the training set are left unchanged. We also conduct separate training runs where we keep 100\% of the training data unchanged to compare the performance of the models trained with modified data.

\subsection{Models}
For fine-tuning, we used pre-trained versions of models that are available on the Huggingface Hub. We perform our experiments on the following four transformer-based architectures:

\begin{enumerate}
    \item \textbf{BERT}: Bidirectional Encoder Representations from Transformers (BERT) model was proposed by the authors of \cite{devlin2018bert}. The BERT model was pre-trained using the masked language modeling (MLM) objective. The objective allows the model to predict the original word using context from both the left and right sides of masked tokens. In addition to MLM, the model was also pre-trained using the Next Sentence Prediction (NSP) objective. We utilized the BERT-Base\footnote{\url{https://huggingface.co/google-bert/bert-base-uncased}} model, which has approximately 110M parameters.
    \item \textbf{DistilBERT}: DistilBERT \cite{sanh2019distilbert} is a distilled version of the BERT model. The DistilBERT model decreases the BERT model's parameters by almost 40\% percent, making it much smaller and faster than the Base BERT model. We utilized the DistilBERT-Base\footnote{\url{https://huggingface.co/distilbert/distilbert-base-uncased}} model, which only has approximately 66M parameters.  
    \item \textbf{RoBERTa}: The RoBERTa \cite{liu2019roberta} model was proposed in 2019 by the researchers at Facebook AI. They observed that BERT was greatly undertrained, and improved pre-training methods can be used to enhance the performance of the original BERT model. They named this new method as the Robustly optimized BERT approach (RoBERTa). We utilized the RoBERTa-Base\footnote{\url{https://huggingface.co/FacebookAI/roberta-base}}, which has approximately 125M parameters.
    \item \textbf{Bio-Discharge Summary BERT}: The authors of \cite{alsentzer2019publicly} publicly released multiple versions of BERT embeddings that were fine-tuned on clinical text. Their work showed that using contextual embedding can improve the model's performance on clinical domain-specific tasks. In our study, we use the Bio+Discharge Summary BERT\footnote{\url{https://huggingface.co/emilyalsentzer/Bio_Discharge_Summary_BERT}} (BDS-BERT) model. BDS-BERT was initialized from BioBERT \cite{lee2020biobert} and fine-tuned on discharge summaries from the MIMIC dataset. The model has an approximate parameter count of 108M.
\end{enumerate}

\subsection{Hyperparameter Settings}
The training was done with the NVIDIA Tesla V100 (16GB) GPU available on Google Colab. We train each model for 10 epochs with a batch size of 16. The maximum input text block size is 512. We use the AdamW optimizer with a learning rate of $2e-5$ and weight decay of $0.01$. In addition, we set the scheduler type as `constant', which keeps the learning rate constant throughout the training period. 

\subsection{Evaluation Metrics} \label{sec:eval_metrics}

\begin{enumerate}
    \item \textbf{Hamming Loss:} The traditional accuracy metric is not suitable for MLC tasks, as it does not account for the fraction of labels that are correctly predicted for a particular sample. For that reason, we used the hamming loss (HL) as an evaluation metric, which is more suitable for MLC problems. The hamming loss indicates the percentage of labels that are incorrectly predicted. It is calculated by dividing the total number of incorrectly predicted labels (TNIPL) by the total number of predicted labels (TNPL). A lower hamming loss score indicates higher model performance. The hamming loss is defined by \textbf{equation \ref{eq_hamml}}:
    
    \begin{equation}
    Hamming\:Loss=\frac{TNIPL}{TNPL}\label{eq_hamml}
    \end{equation}
    
    \item \textbf{Precision:} In our study, the precision (P) score refers to the proportion of the correctly predicted diseases out of all the diseases included in the differential diagnosis predicted by the model. A higher precision score indicates that the model generates fewer false positives (FP). We employed the `samples' average precision score, which is suitable for MLC tasks. The `samples' average calculates the metric for each sample present in the dataset and then determines the average to obtain the final score. The precision score is defined by \textbf{equation \ref{eq_precision}}: 
    
    \begin{equation}
    Precision=\frac{TP}{TP + FP}\label{eq_precision}
    \end{equation}

    Here, true positives (TP) are diseases that are correctly included in the differential diagnosis, and false positives (FP) are diseases that are incorrectly included in the differential diagnosis.
    
    \item \textbf{Recall:} In our study, the recall (R) score is the percentage of diseases from the ground-truth differential diagnosis that was correctly included in the predicted differential diagnosis. A higher recall score indicates that the model generates fewer false negatives (FN). Similar to the precision score, we also used the `samples' average recall score in our evaluations. The recall score is defined by \textbf{equation \ref{eq_recall}}:

    \begin{equation}
    Recall=\frac{TP}{TP + FN}\label{eq_recall}
    \end{equation}

    Here, false negatives (FN) are instances where the model incorrectly excludes diseases from the differential diagnosis.
    
    \item \textbf{F1}: The f1 score is a harmonic mean of the precision and recall scores. It is an essential metric if we want to prioritize both precision and recall scores. The f1 score is defined by \textbf{equation \ref{eq_f1}}:

    \begin{equation}
    F1 =2\times\frac{Precision\times Recall}{Precision+Recall}\label{eq_f1}
    \end{equation}

    \item \textbf{GTD Score:} As mentioned previously, the DDXPlus dataset provides the ground truth pathology (disease) from which the patient is suffering. In order to determine whether the ground truth disease is present in the predicted differential diagnosis, we employ the Ground Truth in Differential (GTD) Score. The GTD score is similar to the GTPA score presented in the DDXPlus paper \cite{fansi2022ddxplus}. Let $D$ be the set of all diseases present in the dataset, and $D_{diff} \subset D$  be a differential diagnosis for a patient as predicted by the model. Also, let $D_{gt} \in D$ be the ground truth disease a patient is suffering from, and $N$ be the total number of samples (patients) in the dataset. The GTD score can be defined by \textbf{equation \ref{eq_gtd}}:
    
    \begin{equation}
    GTD\:Score = \frac{\sum_{i=1}^{N} P^i}{N}\\
    \;,where\:\:P^i = \begin{cases}
           1, & \text{if $D_{gt}^i \in D_{diff}^i$}\\
           0, & \text{otherwise}
          \end{cases}
    \label{eq_gtd}
    \end{equation}

    Here, for the $i^{th}$ patient in the dataset, the value of $P^i$ is set to 1 if the ground truth $D_{gt}^i$ is present in the model predicted differential $D_{diff}^i$. In our work, we included different variations of the GTD score to evaluate our models. For instance, GTD@0.95 was used when the differential was obtained by setting the model confidence threshold $T_{conf}$ (mentioned in \textbf{section \ref{sec:model_evaluation}}) to 0.95. Setting the $T_{conf}$ value to 0.95 shrinks the differential by only including diseases for which the model produced very high confidence scores. Similarly, GTD@top-5 was used to see whether the ground truth disease is present among the top-5 diseases present in the differential. For the default GTD score, the $T_{conf}$ value is set to 0.5.
\end{enumerate}

%%%%%%%%%%%%%%%%%%%%%%%%%%%%%%%%%%%%%%%%%%%%%%%%%%%%%%%%%%%%%%%
\section{Results and Discussion}
As mentioned earlier in \textbf{section \ref{sec:dataset_dist}}, we trained all four models twice: once with modifications applied to the training data and once without modifications. We refer to the models fine-tuned without data modifications as `base' models. In comparison, we refer to the models fine-tuned with data modification approaches such as sequence paraphrasing (SP) and medical term diversification (MTD) as `SP-MTD' models. 

Throughout this section, we will first demonstrate the performance comparison across all models on the held-out test set. Next, we present the results obtained from the behavioral tests described in \textbf{section \ref{sec:behavioral_test}}. Finally, we provide a thorough discussion of our findings from this study.

\subsection{Performance on Held-out Test Set}

\textbf{Table \ref{tab:table_1}} showcases the complete results related to the performance of the models on the held-out test set. All four models demonstrated remarkable performance on the held-out test set, achieving an f1 score of over 97\% in each case. It is also evident from the table that data modification approaches had minimal impact on enhancing model performance on the held-out set. In many instances, the base models demonstrated higher scores, though the difference was often very small (less than 0.5\%). However, this result was not unexpected since the non-modified training set contained more samples that resembled the type of samples present in the held-out test set. In comparison, the training set with modifications had 30\% of its samples modified, and those samples had less resemblance to the samples in the held-out set. Nevertheless, the models trained with modified data did not show any significant decrease in performance.

\begingroup
\renewcommand{\arraystretch}{1.5}

\begin{table*}[b]
\caption{\textbf{Performance comparison of models on the held-out test set.} Bold numeric values indicate the highest scores obtained across all 4 models. Colored numerical values show instances where a model trained with data modifications obtained higher scores than the same model trained without modifications.}
\centering
\resizebox{\linewidth}{!}{
\begin{tabular}{lccccccc}
\hline
\multirow{2}{*}{Models} & \multicolumn{7}{c}{Metrics (\%)} \\
\cmidrule{2-8}
 & HL & F1 & P & R & GTD@0.5 & GTD@0.95 & GTD@top-5 \\
\hline
BERT-base  & \textbf{0.82} & \textbf{97.44} & 97.18 & \textbf{98.24} & \textbf{99.94} & 97.04 & 76.10 \\
 BERT-SP-MTD  & 0.87 & 97.31 & \cellcolor[HTML]{33FFE0} 97.33 & 97.83 & 99.83 & 95.45 & 75.35 \\
\hline
DistilBERT-base  & 0.91 & 97.04 & 96.98 & 97.74 & 99.83 & \textbf{98.26} & 77.03 \\
DistilBERT-SP-MTD & 0.94 & \cellcolor[HTML]{33FFE0} 97.08 & \cellcolor[HTML]{33FFE0} 97.15 & 97.64 & 99.83 & 97.35 & \cellcolor[HTML]{33FFE0} \textbf{77.23} \\
\hline
RoBERTa-base & 0.88 & 97.28 & 97.09 & 98.05 & \textbf{99.94} & 97.11 & 74.69 \\
RoBERTa-SP-MTD & \cellcolor[HTML]{33FFE0} 0.87 & 97.26 & \cellcolor[HTML]{33FFE0} 97.18 & 97.96 & 99.73 & 96.67 & 74.34 \\
\hline
BDS-BERT-base & 0.88 & 97.36 & 97.21 & 98.06 & 99.92 & 95.84 & 74.94 \\
BDS-BERT-SP-MTD & 0.88 & 97.30 & \cellcolor[HTML]{33FFE0} \textbf{97.36} & 97.79 & 99.83 & \cellcolor[HTML]{33FFE0} 95.89 & \cellcolor[HTML]{33FFE0} 75.48 \\
\hline
\end{tabular}
}
\label{tab:table_1}
\end{table*}
\endgroup

Overall, the BERT-base model demonstrated the best performance across the hamming loss, f1, precision, and recall metrics. On the other hand, the DistilBERT models (both base and SP-MTD) recorded the lowest scores, which is expected since DistilBERT is the smallest of the four models. The BERT-base model achieved the best HL score of 0.82, which indicates it incorrectly predicts only 0.82\% of the labels. This model also achieved the highest F1 and recall scores of 97.44 and 98.24. In comparison, the second-highest F1 and recall scores of 97.36 and 98.06 were obtained from the BDS-BERT-base model. The second-highest HL score of 0.87 was achieved by the BERT-SP-MTD and RoBERTa-SP-MTD models. In terms of precision score, the models trained with data modifications obtained higher scores on all four models. This is an interesting pattern, as it suggests the SP-MTD models were more efficient at generating fewer false positives. This means most of the diseases included in the differentials were predicted accurately. The highest precision score of 97.36 was achieved by the BDS-BERT-SP-MTD model. The opposite is observed in the base models, as all of them had higher recall scores. This indicates the base models predicted a higher percentage of diseases from the ground-truth differentials, although this led to some additional false positives.

Lastly, we can observe from the ground truth in differential (GTD) scores that all models were extremely efficient at including the ground truth disease inside the differential. At the confidence threshold value of 0.5, the lowest GTD score of 99.73 was obtained by the RoBERTa-SP-MTD model. This means that 99.73 or higher percentage of times, the models accurately kept the ground-truth disease inside the predicted differential. The highest GTD@0.5 score was achieved by the BERT-base and RoBERTa-base models, which is 99.94. The GTD scores slightly dropped at differentials obtained at a confidence threshold of 0.95. Moreover, the GTD scores dropped more significantly when only the top 5 diseases were included in the differential. Surprisingly, in both these cases, the smaller DistilBERT models demonstrated the best results. The highest and second-highest GTD@95 scores of 98.26 and 97.35 were obtained from the DistilBERT-base and DistilBERT-SP-MTD models, respectively. Similarly, the top two GTD@top-5 scores of 77.23 and 77.03 were obtained from the DistilBERT-SP-MTD and DistilBERT-base models.

\subsection{Performance on Behavioral Tests}
For the behavioral tests, we present an additional GTD score with a confidence threshold of 0.2. As we intentionally made the behavioral tests challenging, the model's confidence scores for predictions were also likely to decrease. Therefore, we use a lower threshold to see how the performance changes.

\begingroup
\renewcommand{\arraystretch}{1.2}

\begin{table*}[h]
\caption{\textbf{Comparison of hamming loss, f1, precision, and recall scores on behavioral Tests.} Bold numeric values indicate the highest scores obtained across all 4 models for a particular test type. Colored numerical values show instances where a model trained with data modifications obtained higher scores than the same model trained without modifications.}
\centering
\begin{tabular}{clcccc}
\hline
\multirow{2}{*}{\makecell{Test Type}} & \multirow{2}{*}{Models} & \multicolumn{4}{c}{Metrics (\%)} \\
\cmidrule{3-6}
 & & HL & F1 & P & R \\
\hline
\multirow{8}{*}{\rotatebox[origin=c]{90}{\makecell{Typo \\ Insertion}}} & 
BERT-base & 1.92 & 93.70 & 95.01 & 94.02 \\
 & BERT-SP-MTD & \cellcolor[HTML]{33FFE0} 1.88 & \cellcolor[HTML]{33FFE0} 93.84 & \cellcolor[HTML]{33FFE0} 95.54 & 93.70\\
\cmidrule{2-6}

& DistilBERT-base & 2.21 & 92.62 & 94.39 & 92.82\\
& DistilBERT-SP-MTD & \cellcolor[HTML]{33FFE0} 2.09 & \cellcolor[HTML]{33FFE0} 92.99 & 94.37 & \cellcolor[HTML]{33FFE0} 93.39\\
\cmidrule{2-6}

& RoBERTa-base & 1.69 & 94.49 & 95.54 & 94.80\\
& RoBERTa-SP-MTD & \cellcolor[HTML]{33FFE0} \textbf{1.56} & \cellcolor[HTML]{33FFE0} \textbf{94.89} & \cellcolor[HTML]{33FFE0} \textbf{96.09} & \cellcolor[HTML]{33FFE0} \textbf{95.01}\\
\cmidrule{2-6}

& BDS-BERT-base & 1.95 & 93.57 & 95.10 & 93.80 \\
& BDS-BERT-SP-MTD & \cellcolor[HTML]{33FFE0} 1.88 & \cellcolor[HTML]{33FFE0} 93.91 & \cellcolor[HTML]{33FFE0} 95.49 & \cellcolor[HTML]{33FFE0} 93.91  \\
\hline

\multirow{8}{*}{\rotatebox[origin=c]{90}{\makecell{Medical History \\ Exclusion}}} & BERT-base & \textbf{1.03} & \textbf{96.89} & 96.63 & \textbf{97.80} \\
 & BERT-SP-MTD & 1.11 & 96.71 & \cellcolor[HTML]{33FFE0} 96.81 & 97.33\\
\cmidrule{2-6}

& DistilBERT-base & 1.16 & 96.49 & 96.50 & 97.24\\
& DistilBERT-SP-MTD & 1.20 & 96.38 & \cellcolor[HTML]{33FFE0} 96.61 & 96.91\\
\cmidrule{2-6}

& RoBERTa-base & 1.09 & 96.73 & 96.49 & 97.64\\
& RoBERTa-SP-MTD & \cellcolor[HTML]{33FFE0} 1.07 & \cellcolor[HTML]{33FFE0} 96.84 & \cellcolor[HTML]{33FFE0} 96.83 & 97.50\\
\cmidrule{2-6}

& BDS-BERT-base & 1.10 & 96.74 & 96.74 & 97.44 \\
& BDS-BERT-SP-MTD & \cellcolor[HTML]{33FFE0} 1.08 & \cellcolor[HTML]{33FFE0} 96.86 & \cellcolor[HTML]{33FFE0} \textbf{96.96} & 97.41  \\
\hline

\multirow{8}{*}{\rotatebox[origin=c]{90}{\makecell{Custom \\ Test Set}}} & BERT-base & 13.18 & 63.66 & 67.27 & 72.72 \\
& BERT-SP-MTD & \cellcolor[HTML]{33FFE0} 10.76 & \cellcolor[HTML]{33FFE0} 69.95 & \cellcolor[HTML]{33FFE0} 69.54 & \cellcolor[HTML]{33FFE0} \textbf{78.54}\\
\cmidrule{2-6}

& DistilBERT-base & 12.12 & 61.14 & \textbf{72.75} & 61.32\\
& DistilBERT-SP-MTD & \cellcolor[HTML]{33FFE0} \textbf{10.16} & \cellcolor[HTML]{33FFE0} 68.81 & 72.44 & \cellcolor[HTML]{33FFE0} 72.78\\
\cmidrule{2-6}

& RoBERTa-base & 11.18 & 65.32 & 70.63 & 69.54\\
& RoBERTa-SP-MTD & \cellcolor[HTML]{33FFE0} 10.18 & \cellcolor[HTML]{33FFE0} 68.42 & 68.54 & \cellcolor[HTML]{33FFE0} 74.80\\
\cmidrule{2-6}

& BDS-BERT-base & 11.55 & 64.70 & 71.72 & 69.58 \\
& BDS-BERT-SP-MTD & \cellcolor[HTML]{33FFE0} 10.55 & \cellcolor[HTML]{33FFE0} \textbf{70.11} & 71.48 & \cellcolor[HTML]{33FFE0} 77.13  \\
\hline
\end{tabular}
\label{tab:table_2}
\end{table*}

\endgroup

\subsubsection{Test Set with Typo Insertion}
In terms of HL score, \textbf{table \ref{tab:table_2}} shows that all models experienced a slight increase in error rate when typos were added to the test set. Both BERT and BDS-BERT type models experienced around a 1\% increase in HL score. We can also see a 3-4\% decrease in f1, precision, and recall scores for all BERT and BDS-BERT models. As DistilBERT was the smallest model, both DistilBERT and DistilBERT-SP-MTD were most susceptible to the typos inserted in the test set. Both of these models achieved the lowest scores across all four metrics. On the other hand, both the RoBERTa-base and RoBERTa-SP-MTD models attained the best scores on the typo-inserted test set. The RoBERTa-SP-MTD models achieved the highest HL, f1, precision, and recall scores of 1.56, 94.89, 96.09, and 95.81, respectively. Overall, considering the four evaluation metrics in \textbf{table \ref{tab:table_2}}, the SP-MTD models had a slight edge over the base models in most cases.

\begin{figure*}
    \centering         
    \includegraphics[width=\textwidth]{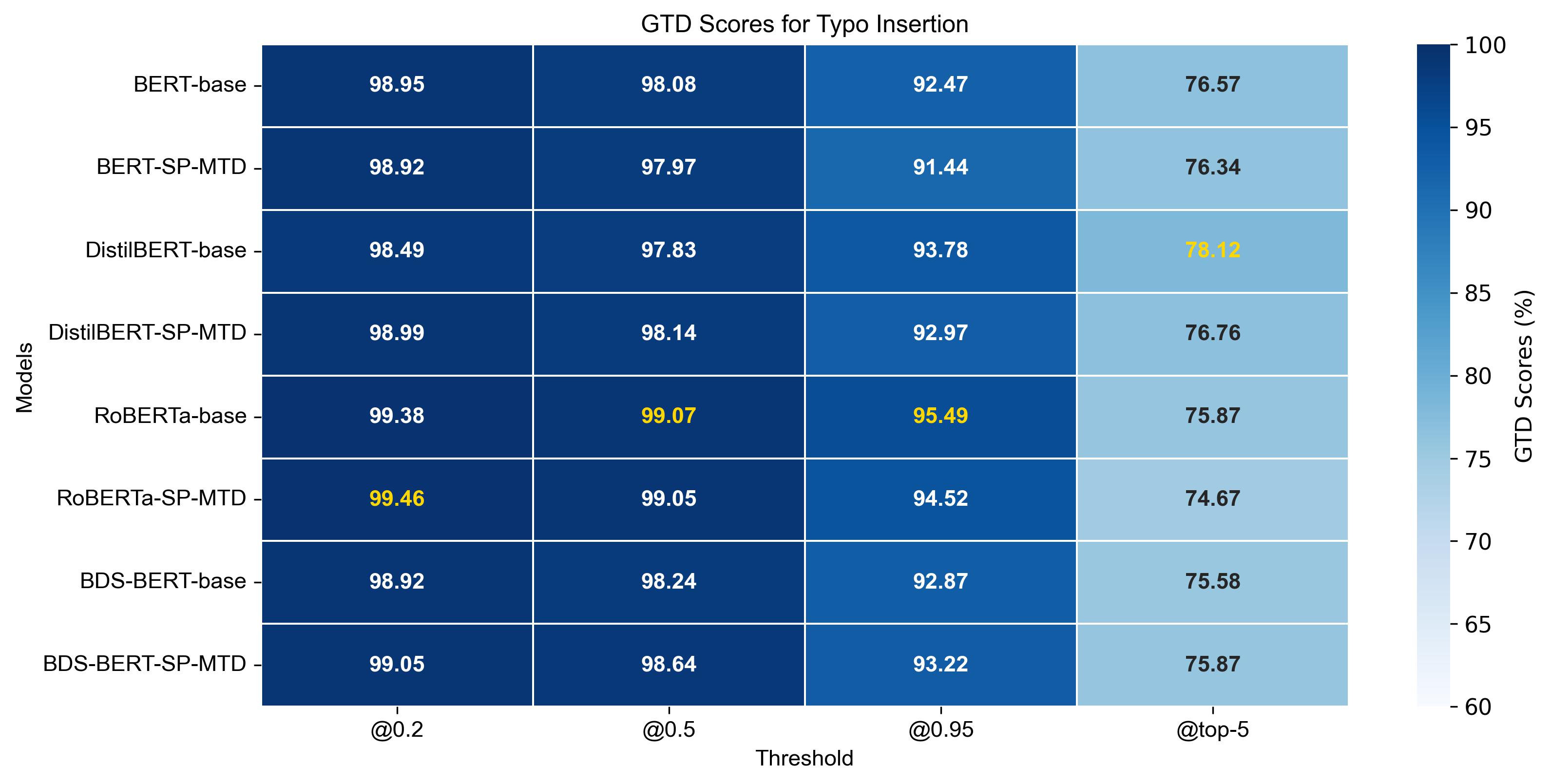}
    \caption{\textbf{Comparison of ground truth in differential (GTD) scores for the typo insertion test.} The highlighted (in yellow) numeric values indicate the highest scores obtained across all models for a particular threshold.}
    \label{fig:GTD_typo}
\end{figure*}

The results in \textbf{Fig \ref{fig:GTD_typo}} show that the GTD scores were slightly affected as typos were introduced into the test set. The RoBERTa models were again the least susceptible to the typo-introduced test set. At threshold 0.5 (which we consider the default threshold), the best GTD@0.5 score of 99.07 was obtained by the RoBERTa-base model. Previously, the same model obtained a GTD@0.5 score of 99.94. The drop in GTD scores was a bit higher for other models. For example, for the BERT-base model, the GTD@0.5 score dropped from 99.94 to 98.08. Similarly, the DistilBERT-base model's GTD@0.5 score dropped from 99.83 to 97.83. At the 0.95 threshold, the RoBERTa-base model again achieved the highest GTD@0.95 score of 95.49. The smaller DistilBERT models demonstrated a much larger drop at this threshold. For example, previously, the DistilBERT-base model achieved the highest GTD@0.95 score of 98.26 on the held-out test set. However, this score dropped to 93.78 when typos were introduced. However, for the top-5 predictions, the DistilBERT-base model still achieved the highest GTD@top-5 score of 78.12, which is actually higher than the GTD@0.95 score obtained on the held-out set. It is worth noting that these GTD scores can be improved by lowering the threshold to 0.2, as evident from \textbf{Fig \ref{fig:GTD_typo}}. The RoBERTa-SP-MTD model achieved the best GTD@0.2 score of 99.46.

\subsubsection{Test Set with Medical History Exclusion}
\textbf{Table \ref{tab:table_2}} shows that randomly excluding medical history information from the test set had the least effect on the model's performance compared to other behavioral tests. Considering the HL score, we can see that the error rate for the BERT-base model increased to 1.03. In comparison, this model obtained the best HL score of 0.82 on the held-out test set. All models exhibited similar rates of increase in error, which were only in the range of 0.2\% to 0.3\%. We also recorded a small decrease in the f1, precision, and recall scores. However, these decreases were also very minimal, less than 1\%. Similar to the held-out test set, the BERT-base model attained the highest HL, f1, and recall scores of 1.03, 96.89, and 97.80, respectively. On the other hand, the SP-MTD models achieved better precision scores, with the BDS-BERT-SP-MTD model achieving the highest score of 96.96. This phenomenon was also observed in the case of the held-out test set.

\begin{figure*}
    \centering         
    \includegraphics[width=\textwidth]{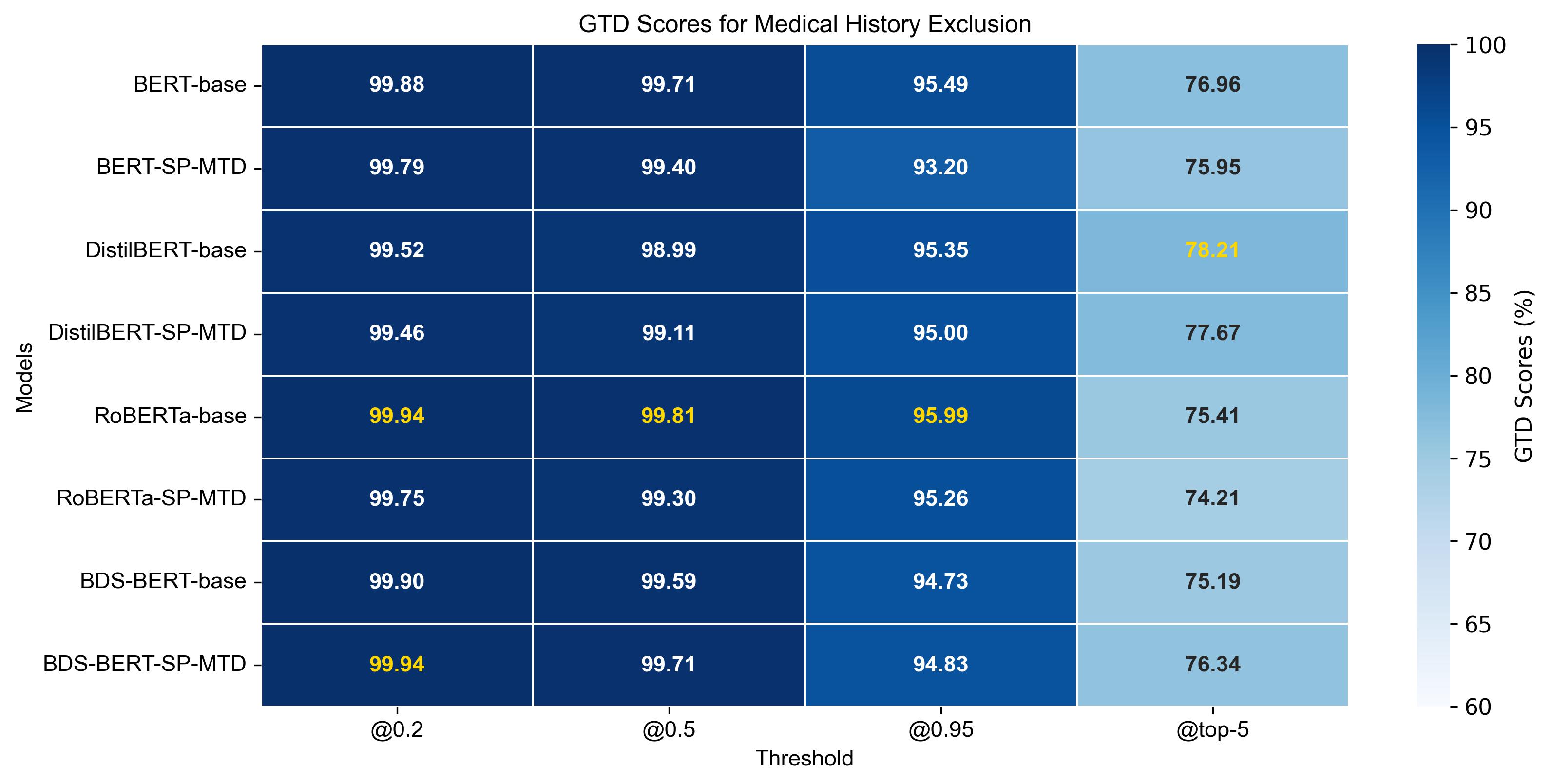}
    \caption{\textbf{Comparison of ground truth in differential (GTD) scores for the medical history exclusion test.} The highlighted (in yellow) numeric values indicate the highest scores obtained across all models for a particular threshold.}
    \label{fig:GTD_mhe}
\end{figure*}

Additionally, \textbf{Fig \ref{fig:GTD_mhe}} shows that this behavioral test did not significantly impact the model's GTD scores either. Particularly the GTD scores at the default threshold of 0.5. In most cases, the decrease was less than 0.5\%. The RoBERTa-base model held the highest GTD@0.5 score of 99.81. We do notice a slightly larger decrease when the confidence threshold is increased to 0.95. The RoBERTa-base model also achieved the best GTD@0.95 score of 95.99. Out of the four models, the GTD scores of the DistilBERT models were affected the most at thresholds 0.5 and 0.95. However, for the top-5 predictions, the DistilBERT models still achieved better results compared to others, with the best two GTD@top-5 scores of 78.21 and 77.67. Like the previous behavioral test, the GTD scores can be improved by decreasing the threshold to 0.2. For example, the RoBERTa-base and the BERT-SP-MTD models achieved the best GTD@0.2 score of 99.94. This score matches the previous best score obtained for the original held-out test set at the 0.5 threshold.

\subsubsection{Custom Test Set}
This behavior test was designed to be the most challenging among the three test cases. The results obtained from this custom test set of 100 samples further demonstrated the difficulty level of the test set. From \textbf{table \ref{tab:table_2}}, we can see a significant decrease across all evaluation metric scores. In particular, the base models were most affected by the perturbations introduced in the custom test set. However, out of the four base models, the RoBERTa and BDS-BERT base models were least susceptible to the custom test set. In contrast, the SP-MTD models demonstrated significant performance improvements over the base model in almost all cases. 

In \textbf{table \ref{tab:table_2}}, we can see from the HL scores that the error rate increased to over 10\% for all of the models. In particular, the BERT-base model obtained the lowest HL score of 13.18. This indicates that the model incorrectly predicts 13.18\% of the labels. In comparison, the DistilBERT-SP-MTD and RoBERTa-SP-MTD models obtained the highest HL scores of 10.16 and 10.18, respectively. In terms of the F1 score, the SP-MTD models demonstrated a 3-7\% improvement over the base models. The highest F1 score of 70.11 was obtained by the BDS-BERT-SP-MTD model. The second-highest F1 score was obtained by the BERT-SP-MTD model, which was 69.95. 

We also found an interesting finding by looking at the precision and recall scores. For three of the models, we can see the base models seem to have a slightly higher precision when compared to the SP-MTD model. However, when considering the recall scores, we can see that the SP-MTD models improved performance significantly (by 5\%-11\%) compared to the base models. We can deduce from this observation that the SP-MTD models were able to accurately predict a higher percentage of diseases from the ground truth differential when compared to the base models. In simple terms, it means that the SP-MTD models were generating better differential diagnoses, which sometimes included a few false positives. In comparison, the base models generated smaller differentials, leading to a decrease in the recall scores as they excluded a lot of diseases from the ground-truth differentials. As smaller differentials mean fewer false positives, it resulted in a slightly higher precision score. For example, the DistilBERT-base model obtained the highest precision score of 72.75; however, it got the lowest recall score of 61.32. In comparison, the modifications applied during the fine-tuning of DistilBERT-SP-MTD resulted in it achieving a recall of 72.78, which is an 11\% improvement over the base model. On top of that, the DistilBERT-SP-MTD model obtained the second-highest precision score of 72.75. Overall, we can see a significant increase in recall scores for all SP-MTD models, with the BERT-SP-MTD and the BDS-BERT-SP-MTD models obtaining the highest scores of 78.54 and 77.13, respectively.

\begin{figure*}
    \centering         
    \includegraphics[width=\textwidth]{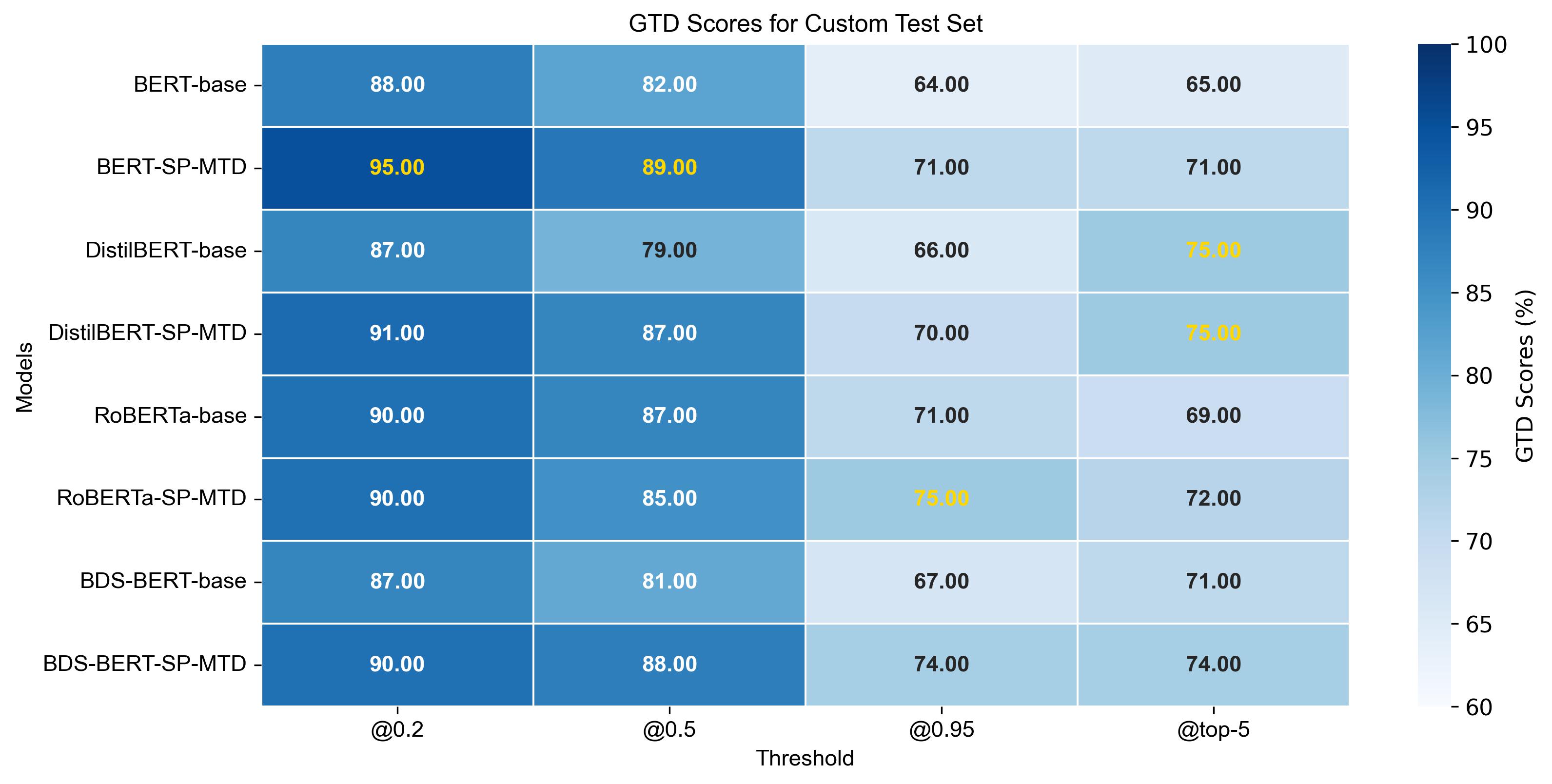}
    \caption{\textbf{Comparison of ground truth in differential (GTD) scores for the custom test set.} The highlighted (in yellow) numeric values indicate the highest scores obtained across all models for a particular threshold.}
    \label{fig:GTD_custom_set}
\end{figure*}

As we had exactly 100 samples in the custom test set, the GTD scores show exactly how many instances out of the 100 the models correctly kept the ground-truth diagnosis within the differential. We can observe one thing right away by looking at \textbf{Fig \ref{fig:GTD_custom_set}}, which is the GTD scores were not as badly affected as the other evaluation metrics for the custom test set. This means even though the models did suffer in making highly accurate differentials, they still managed to identify the ground truth disease. At threshold 0.2, the BERT-SP-MTD model correctly included the ground-truth disease 95 times out of 100. This was the best score out of all the models. The second-highest GTD@0.2 score was obtained by the DistilBERT-SP-MTD model, which was 91. At threshold 0.5,  the BERT-SP-MTD model again obtained the highest GTD@0.5 score of 89. The BDS-BERT model achieved the second-best score of 88. At threshold 0.95, the two highest GTD@0.95 scores of 75 and 74 were obtained by the RoBERTa-SP-MTD and BDS-BERT-SP-MTD, respectively. When considering the top-5 predictions, the DistilBERT model again showed superiority over other models, achieving a GTD@top-5 score of 75 on both the base and SP-MTD models. Overall, similar to the other evaluation metrics, the SP-MTD demonstrated better GTD scores on the custom test set than the base models.

\subsection{Discussion}
\begin{figure*}
    \centering         
    \includegraphics[width=\textwidth]{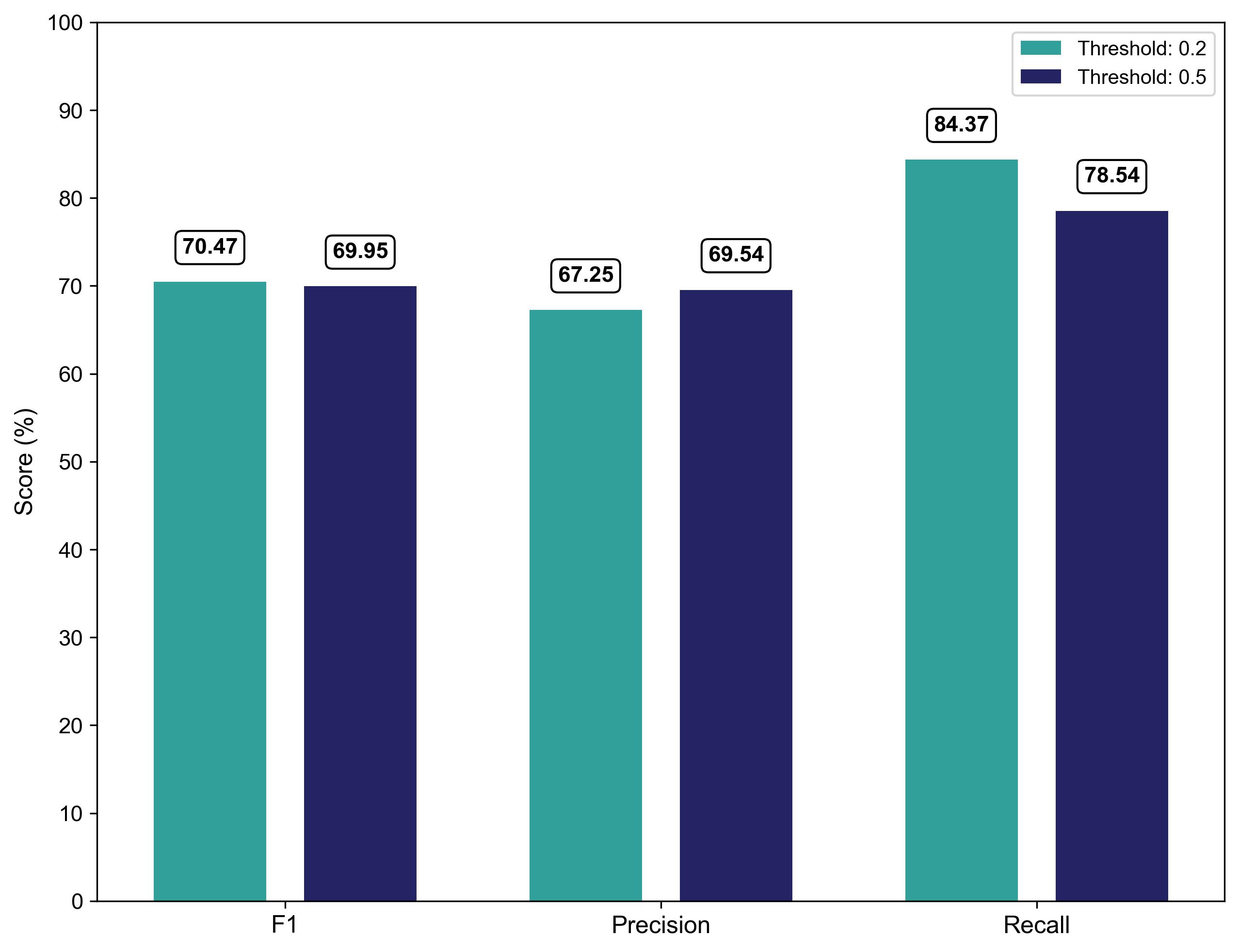}
    \caption{\textbf{Comparison of F1, precision, and recall metrics for the custom test set.} These scores were obtained from the BERT-SP-MTD model at confidence thresholds 0.2 and 0.5.}
    \label{fig:metrics_at_thresholds}
\end{figure*}

Our study shows promising results on the potential application of smaller encoder-based transformer models for performing differential diagnosis. All four models used in our study yield impressive results on the held-out test set. In addition, we put our models through rigorous testing to identify the potential limitations that can be addressed in future research. Overall, our findings from the behavioral tests can be summarized as follows:

\begin{enumerate}
    \item  When considering the base models, the performance of the RoBERTa-base model was least affected by the typo insertion test and the custom test set. The advanced optimization techniques applied during the pre-training of RoBERTa may have contributed to its superior performance. Additionally, the BDS-BERT-SP-MTD model achieved the best F1 score on the Custom test set. The medical data used for pre-training the BDS-BERT model may have been the reason why the model performed well on such challenging samples.

    \item The behavioral test also showed that the perturbations affected the GTD scores less when compared to other evaluation metrics. The results also showed that decreasing the threshold value can improve the GTD scores. However, the threshold value should be adjusted carefully. The reason is that lowering the threshold value will naturally increase the differential size, which can result in some false positives. For example, \textbf{Fig \ref{fig:metrics_at_thresholds}} shows the F1, precision, and recall scores for the custom set at thresholds 0.2 and 0.5, which were obtained from the BERT-SP-MTD model. It shows that lowering the threshold to 0.2 improved the F1 score. However, we can see that the precision score decreased when compared to the default threshold of 0.5. This occurs due to an increase in the number of false positives. In contrast, the recall score improved significantly at a lower threshold due to larger differentials, which ultimately increased the overall F1 score. Therefore, the threshold value of the model's confidence score needs to be adjusted based on the requirements of its user.
    
    \item The fine-tuned models were susceptible to typos within the input text to a certain extent. We introduced typos to 50\% of the sentences within each sample. Even with such high numbers of typos, it is important to note that the model's performance dropped by only 2-4\% when considering the F1, precision, and recall metrics. However, it is worth investigating further in future studies by developing strategies to fine-tune these models to be less sensitive to such changes.

    \item The medical history exclusion test revealed that the models were not heavily dependent on the medical history of the patient to make an accurate differential diagnosis. Even with some extreme cases where the medical history information was completely removed from the samples, the model's performance dropped by a very small margin. This indicates that the models were more dependent on the patient's symptoms information when making a diagnosis. This is a positive outcome because sometimes there might be missing information about a patient's medical history.

    \item One of the most important findings of our research is the results obtained on the custom test set. This paves the way for future research into how we can tackle the problem of reduced performance when the input text contains heavy perturbations. Our data modification approaches were able to help the models have a better contextual understanding and improved their performance on the custom test set when compared to the base models. Future research can address this issue by implementing more advanced data processing and fine-tuning methods. 
    
    \item Lastly, the behavioral test with the custom test set revealed the need for newer and more advanced differential diagnosis datasets that are based on real-world data. Our work was limited by the lack of available datasets that provide differential diagnosis information. We had to go through a number of data processing steps to make the DDXPlus dataset suitable for our research goals. However, even with such limitations, our fine-tuned models still showed promising results on the custom test set. Therefore, creating new datasets that include patient records with differential diagnoses covering a wide range of medical conditions would definitely result in robust diagnostic models.
\end{enumerate}

The findings of our study demonstrate that transformer-based architectures can be a useful tool in the healthcare industry for performing the task of differential diagnosis. These types of models can be used to design AI-assistant tools for healthcare professionals. In particular, junior physicians with less experience may find these resources to be of great use. Doctors could utilize these tools to verify their own differential diagnosis. Moreover, these tools can also assist doctors in identifying diseases that may have been overlooked in the initial differential diagnosis. 

Furthermore, AI-assisted tools for differential diagnosis can improve the overall process of developing automated diagnosis systems. Many researchers have already been working on developing image and text analysis approaches for providing disease diagnosis based on results from medical tests. These types of technology can be combined with automatic differential diagnosis tools to design multi-modal diagnostic systems. These systems could first provide a differential diagnosis based on the patient's symptoms and recommend medical tests. The results obtained from these medical tests could be further analyzed by the diagnostic system to provide the final disease diagnosis. However, it is true that developing such reliable systems will take tremendous resources and years of research. We hope that future researchers will be able to use our study as a stepping stone toward developing reliable differential diagnosis tools.

\section{Conclusion}
There is a significant lack of AI-driven research that focuses on performing the task of differential diagnosis. In this study, we trained encoder-based transformer architectures with the goal of providing differential diagnoses based primarily on the patient's symptoms and medical history. We approached this task as a multi-label sequence classification problem. We took patient samples from the DDXPlus dataset and engineered them to construct patient reports that were suitable for the goals of our study. These samples were then fine-tuned using four different encoder-based transformer models. Furthermore, we proposed two methods for applying modifications to the training data in order to improve the model's robustness. 

Through rigorous experiments, we empirically showed that transformer-based models have the potential to be used for developing diagnostic systems that can help with providing differential diagnoses. All the selected models performed remarkably on the held-out test set, achieving F1 scores of over 97\%. In addition to the held-out test set, we also developed behavioral test cases to gain further insight into the capabilities of our models. In particular, we worked with a medical doctor to prepare a new and challenging custom test set. Models trained with our proposed data modification demonstrated superior performance on this custom test set, with one of the models achieving the best F1 score of 70.11\%. 

Overall, the behavioral tests provided us with valuable information regarding some of the limitations of the models that can be addressed in future works. Further research is needed to develop methods that can improve the model's generalization capabilities. Future researchers could also design newer types of behavioral tests relevant to the automatic differential diagnosis tasks. This will help us develop even more reliable systems. Most importantly, more research needs to be done to create large-scale text datasets that contain the differential diagnosis information for patients. We hope our work will inspire other researchers to explore this often-overlooked domain of healthcare NLP.

\section*{Acknowledgments}
This study was funded by North South University, Dhaka, Bangladesh.

\section*{Data Availability Statement}
The source code for data processing, model training, evaluation, and behavioral testing will be made publicly available upon acceptance of our paper. Additionally, the trained models and the custom test set will also be released at that time. However, a demo of our project is publicly available on Huggingface Spaces at: \url{https://huggingface.co/spaces/AdnanSadi/Differential-Diagnosis-Tool}, where all eight models can be accessed for inference.

The DDXPlus dataset used in this study is publicly available at the following repository: \url{https://figshare.com/articles/dataset/DDXPlus_Dataset/20043374}.

\medskip
\bibliography{bibliography}

%%% END INSTRUCTIONS %%%

\end{document}